\pdfoutput=1

\documentclass[11pt]{article}

\usepackage[preprint]{acl}

\usepackage{arydshln}
\usepackage{times}
\usepackage{adjustbox}
\usepackage{amsmath}
\usepackage{latexsym}
\usepackage{multirow}
\usepackage{booktabs}
\usepackage{makecell}

\usepackage{xcolor}
\usepackage{colortbl}
\usepackage[utf8]{inputenc}

\usepackage{microtype}

\usepackage{inconsolata}
\usepackage{hyperref}
\usepackage{cleveref}

\Crefname{section}{§}{§§}
\Crefname{section}{§}{§§}

\usepackage{enumitem}

\usepackage{graphicx}

\title{Delusions of Large Language Models}

\author{Hongshen Xu\textsuperscript{1}\footnotemark[1], Zixv yang\textsuperscript{1}\footnotemark[1], Zichen Zhu\textsuperscript{1}, Kunyao Lan\textsuperscript{1}, Zihan Wang\textsuperscript{1},\\ \textbf{Mengyue Wu\textsuperscript{1}\footnotemark[2], Ziwei Ji\textsuperscript{2}, Lu Chen\textsuperscript{1}, Pascale Fung\textsuperscript{2}, Kai Yu\textsuperscript{1}\footnotemark[2]}\\
  \textsuperscript{1}X-LANCE Lab, Department of Computer Science and Engineering \\
  MoE Key Lab of Artificial Intelligence, AI Institute\\
  Shanghai Jiao Tong University, Shanghai, China 
  \\
 \textsuperscript{2}Center for Artificial Intelligence Research (CAiRE),\\Hong Kong University of Science and Technology\\
  \texttt{\{xuhongshen, mengyuewu, kai.yu\}@sjtu.edu.cn} \\}

\begin{document}
\maketitle

\renewcommand{\thefootnote}{\fnsymbol{footnote}}
\footnotetext[1]{Equal contributions.}
\footnotetext[2]{The corresponding authors are Mengyue Wu and Kai Yu.}

\begin{abstract}
Large Language Models (LLMs) often generate factually incorrect but plausible outputs, known as hallucinations. We identify a more insidious phenomenon, LLM delusion, defined as high-belief hallucinations—incorrect outputs with abnormally high confidence, making them harder to detect and mitigate. Unlike ordinary hallucinations, delusions persist with low uncertainty, posing significant challenges to model reliability. Through empirical analysis across different model families and sizes on several Question-Answering tasks, we show that delusions are prevalent and distinct from hallucinations. LLMs exhibit lower honesty with delusions, which are harder to override via fine-tuning or self-reflection. We link delusion formation with training dynamics and dataset noise and explore mitigation strategies such as retrieval-augmented generation and multi-agent debating to mitigate delusions. By systematically investigating the nature, prevalence, and mitigation of LLM delusions, our study provides insights into the underlying causes of this phenomenon and outlines future directions for improving model reliability. 
\end{abstract}
\section{Introduction}

Large Language Models (LLMs, \citealp{dubey2024llama, yang2024qwen2,openai2023gpt4}) have demonstrated remarkable capabilities in natural language understanding and generation, enabling significant advancements across various domains, such as machine translation~\cite{zhang2023prompting}, conversational agents~\cite{yi2024survey}, code generation~\cite{shinn2024reflexion}, etc. These models, trained on vast corpora of text, leverage deep neural architectures to capture intricate linguistic patterns and world knowledge. However, despite their impressive performance, LLMs often suffer from critical limitations, particularly in generating factually incorrect yet plausible-sounding outputs, commonly referred to as hallucinations~\cite{10.1145/3571730}.

\begin{figure}[t]
    \centering
\includegraphics[width=0.9\linewidth]{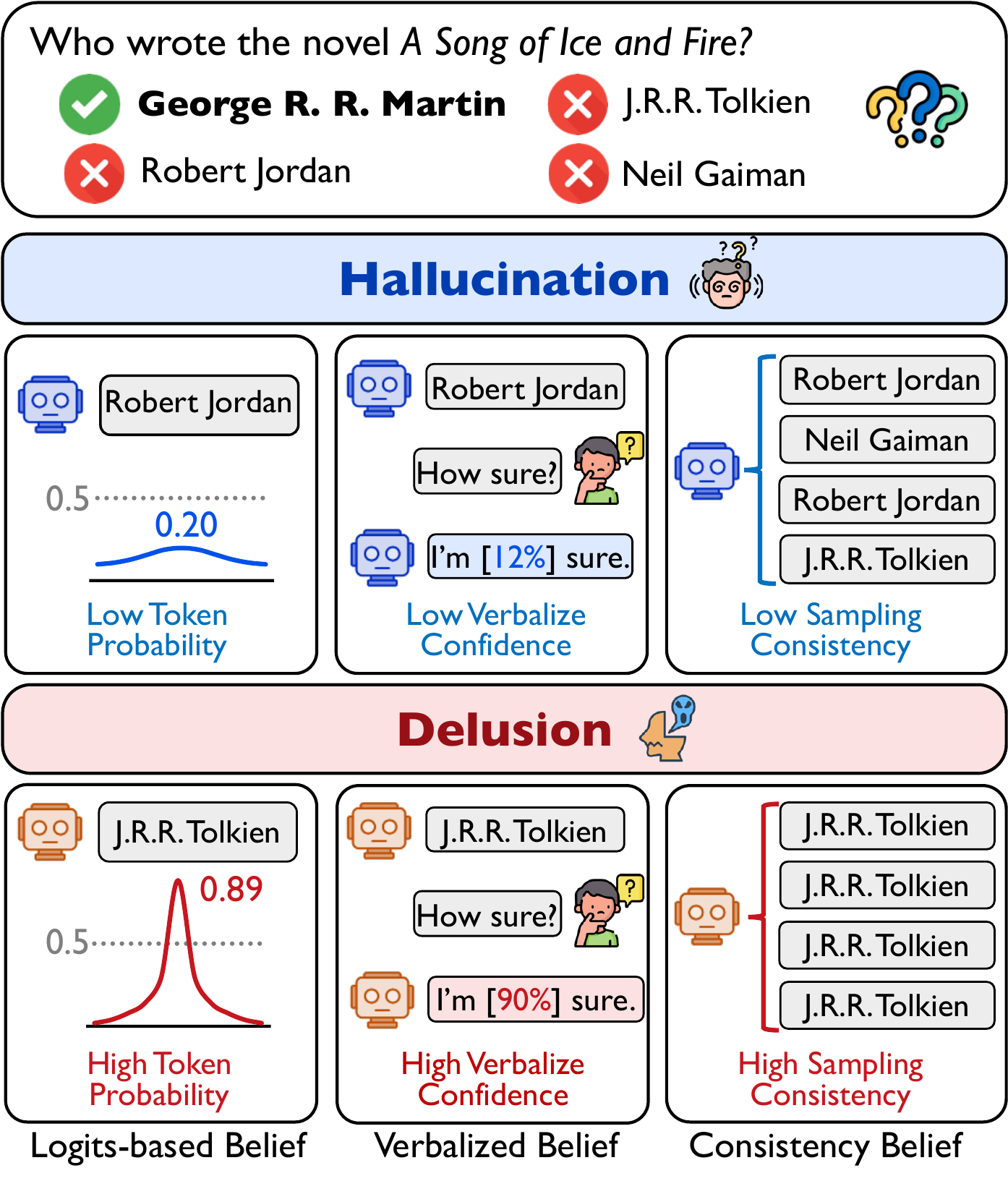}
    \vspace{-0.3cm}
    \caption{Comparative diagram between delusion and hallucination under different belief estimation methods.}
    \label{fig:vs}
    \vspace{-0.5cm}
\end{figure}

While hallucinations in LLMs have been widely studied as incorrect, unfaithful, or nonsensical outputs, we observe that LLMs exhibit an abnormally high level of belief in a subset of hallucinations across different belief probing methods, as shown in Figure 1. Inspired by the concept of delusion~\cite{Kiran2009UnderstandingD} in psychiatry, we term this phenomenon as   \textbf{LLM delusion} for the first time. In psychiatry, delusion is defined as a clearly false belief held with extraordinary conviction that indicates an abnormality in the person’s
content of thought. Similarly, we formally define \textbf{\textit{LLM delusions as high-belief hallucinations}}. Unlike ordinary hallucinations, which often involve high uncertainty and can be flagged through confidence estimation, delusions exhibit low uncertainty, making them particularly difficult to detect and mitigate. This key distinction highlights delusion as a more insidious and persistent challenge in LLMs.

To systematically analyze LLM delusion, we employ uncertainty estimation~\cite{huang2024survey} as a proxy for model belief, using three complementary approaches: (1) \textit{logit-based methods}, which assess token probability distributions; (2) \textit{verbalized confidence}, where the model explicitly states its belief; and (3) \textit{consistency-based methods}, which evaluate belief stability through multiple sampling. To quantify delusion, we introduce a belief threshold, empirically determined as the average confidence assigned to correct answers on a given dataset. If an incorrect response's belief exceeds this threshold, it is classified as delusion. 

In this work, we conduct an extensive empirical investigation into delusions across multiple LLM families and knowledge-intensive Question-Answering benchmarks. Our analysis reveals that delusions are prevalent across different models and sizes, persisting despite variations in uncertainty estimation methods (\Cref{sec:distribution_delusion}). We further designed a series of comparison experiments between delusion and hallucination, revealing that LLM delusions exhibit characteristics similar to psychiatric delusions (\Cref{sec:comparison_delusion}). Additionally, we analyze the formation and dynamics of delusion from both data and training perspectives (\Cref{sec:dynamics}). Finally, we explore several potential mitigation strategies by introducing external verification (\Cref{sec:mitigating}). Our key findings include:
\begin{itemize}[itemsep=0.5pt, topsep=1pt, parsep=0pt]
    \item \textbf{LLMs demonstrate lower honesty with delusions compared to hallucinations.} When prompted to reject unknown knowledge, models are more inclined to refuse standard hallucinations while maintaining delusions.
    \item \textbf{Delusions are significantly harder to override through fine-tuning.}  Even after training LLMs to reject incorrect answers, delusions persist at a higher rate than hallucinations.
    \item \textbf{Self-reflection mechanisms are ineffective at mitigating delusions.} When prompted to reconsider prior responses, LLMs exhibit a strong tendency to reaffirm delusional outputs rather than revising them.
    \item \textbf{The formation of delusions is influenced by both training dynamics and dataset noise.} Through experiments on synthetic datasets, we find that both the proportion and consistency of erroneous information in training data exacerbate delusional tendencies.
    \item External verification methods, such as retrieval-augmented generation and multi-agent debate systems, offer potential pathways for reducing delusions, but significant challenges remain in fully eliminating them.
\end{itemize}

By systematically investigating the nature, prevalence, and mitigation of delusions in LLMs, our study provides insights into the underlying causes of this phenomenon and outlines future directions for improving model reliability. Our findings highlight the necessity of robust verification mechanisms and adaptive confidence calibration to ensure that LLMs can be deployed in real-world applications with greater trustworthiness.

\section{Related Works}
\subsection{Delusion}
Delusion has long been a topic of interest in psychiatry~\cite{Kiran2009UnderstandingD,mourguescodern2024emergencedynamicsdelusionshallucinations}, where it is defined as a belief held with extraordinary subjective certainty, resistant to contrary evidence, and often impossible in content. A key characteristic of delusions is the absolute certainty with which they are maintained, even in the face of overwhelming contradictory evidence. Recently, delusion has gained attention in the field of reinforcement learning~(RL, ~\citealp{zhao2024identifyingaddressingdelusionstargetdirected}), where a candidate target generator and a target estimator are used to simulate belief formation and evaluation processes. Misalignment between these components parallels the mechanisms underlying delusions in the human brain. However, the phenomenon of delusion in natural language generation (NLG) models remains underexplored. Unlike delusion in RL, which typically arises in out-of-distribution scenarios~\cite{pmlr-v162-langosco22a}, delusion in NLG models primarily involves incorrect factual information in real-world contexts. 

\subsection{Uncertainty Estimation}

Uncertainty estimation plays a critical role in evaluating LLM confidence and addressing hallucinations. Traditional approaches, such as Bayesian methods~\cite{shridhar2019comprehensiveguidebayesianconvolutional} and ensemble techniques~\cite{fadeeva-etal-2023-lm,10.5555/3295222.3295387}, have been widely explored but are often computationally expensive and difficult to scale. More recent research leverage methods such as logit-based uncertainty (analyzing token probability distributions,~\citealp{kadavath2022languagemodelsmostlyknow,kuhn2023semantic,DBLP:journals/corr/abs-2402-12348,pmlr-v216-wimmer23a}), consistency-based uncertainty (examining response stability across multiple samples,~\citealp{huang2025unlockingpowerllmuncertainty,wang2023selfconsistency}), and verbalized confidence (explicit model self-assessment,~\citealp{lin2022teachingmodelsexpressuncertainty,tian-etal-2023-just,xiong2024can,NEURIPS2022_8bb0d291,groot-valdenegro-toro-2024-overconfidence}). While these techniques improve hallucination detection, they are not foolproof—models can still generate erroneous outputs with misleadingly low uncertainty. This limitation motivates our investigation into delusion, a phenomenon where models exhibit high confidence in false claims, resisting traditional uncertainty-based filtering.
\subsection{LLM Hallucination}

Hallucination in LLMs refers to generating unfaithful or factually incorrect content~\cite{10.1145/3571730}, posing challenges to reliability in applications such as question answering and knowledge retrieval~\cite{kaddour2023challengesapplicationslargelanguage,pal-etal-2023-med}. 
Extensive efforts have been made to mitigate hallucinations, including factuality-enhanced training~\cite{akyurek-etal-2022-towards,lin-etal-2022-truthfulqa,10.5555/3495724.3495883}, retrieval-augmented generation (RAG)~\cite{gao2024retrievalaugmentedgenerationlargelanguage}, confidence calibration~\cite{huang2024surveyuncertaintyestimationllms}, and reliability alignment~\cite{xu2024rejection, xu2024reducing, zheng2025enhancing}. Our work identifies a more problematic class of hallucination—delusion. Unlike standard hallucinations, delusions persist despite exposure to counterevidence and are harder to eliminate, requiring novel approaches for detection and mitigation.
\section{Delusion in Large Language Models}
\subsection{Definition of Delusion}

\begin{figure}
    \centering
    \includegraphics[width=\linewidth]{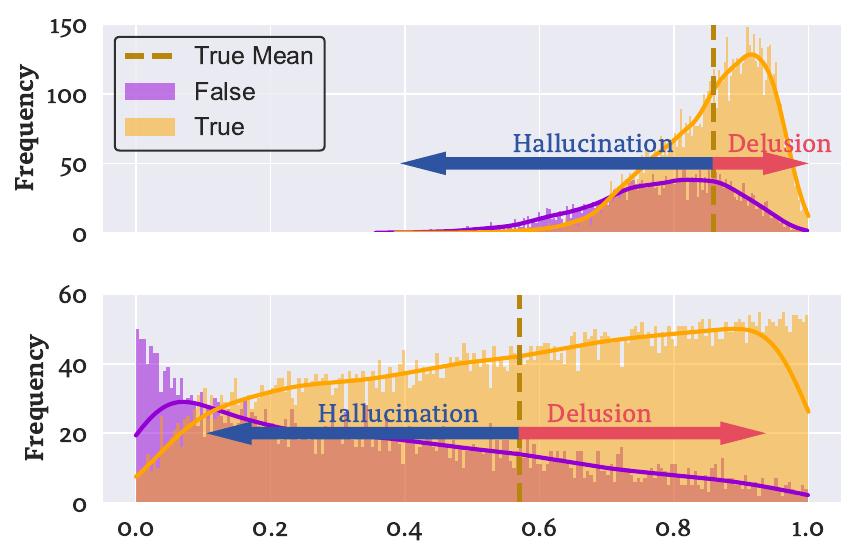}
    \vspace{-0.6cm}
    \caption{Distribution of logits-based belief of Llama3.1-8B-Instruct on TriviaQA test set (with normalized confidence in the lower part).}
    \vspace{-0.5cm}
    \label{fig:conf_distr}
\end{figure}

LLMs exhibit a phenomenon wherein they generate incorrect factual information while simultaneously maintaining a high degree of belief in these inaccuracies. This unwavering confidence persists even when the model is prompted to reassess or confirm its responses. Drawing an analogy from psychiatry~\cite{Kiran2009UnderstandingD}, where delusions refer to strongly held false beliefs, we introduce the concept of delusion in LLMs as a systematic extension of hallucinations. Specifically, we define \textbf{delusion as high-belief hallucinations}, where the model exhibits an anomalously strong conviction in its erroneous outputs. As shown in Figure~\ref{fig:conf_distr}, we classify all false predictions into ordinary hallucinations and delusions based on the estimated belief scores and a specific belief threshold. We will discuss the estimation of \textbf{LLM belief} and \textbf{belief threshold} in the following two subsections.

\subsubsection{Uncertainty as LLM Belief}

Uncertainty estimation has been widely employed in machine learning as a means to assess the confidence of a model in its predictions, providing a measure of the reliability of the model's outputs. Thus we adopt uncertainty estimation as a proxy for the model’s belief in its responses. We employ three primary uncertainty estimation methods:

\noindent \textbf{Logits-based estimation.} This method derives the model’s belief from the probability distribution of output tokens, where a higher probability assigned to a token indicates a stronger belief in that token's correctness. Specifically, we use the \texttt{raw logits}~\cite{lyu2024calibrating} and directly consider the probability of the generation as the confidence. 

\noindent \textbf{Consistency-based estimation.} This method assesses belief by evaluating the stability of the model’s responses over multiple sampling iterations. A response that remains consistent across multiple trials suggests strong belief, whereas variability indicates uncertainty. In this case, we use the answer voting \texttt{agreement}~\cite{lyu2024calibrating} as a measure of consistency.

\noindent \textbf{Verbalized confidence.} Here, the model is explicitly prompted to verbalize its confidence in its response, providing a self-reported measure of belief. We employ three methods to quantify verbalized confidence: \texttt{P(true)}~\cite{kadavath2022languagemodelsmostlyknow}, \texttt{verb. 1S top-1}~\cite{tian2023just}, and \texttt{verb. 2S top-1}~\cite{tian2023just}.

These methods collectively provide a comprehensive understanding of the model’s belief system, recognizing that human belief also varies across different contexts. For the details of uncertainty calculation methods, please refer to Appendix~\ref{sec:uncertainty}.

\noindent \textbf{Normalization of uncertainty scores.} Due to the significant differences in the distributions of uncertainty scores calculated by each method, we also experimented with normalizing these uncertainty scores as the model's belief. This normalization allows for easier comparison and averaging of uncertainty scores derived from different methods. We intuitively normalize the uncertainty scores based on their rankings across all test data. Higher uncertainty scores lead to a higher model rank and a higher normalized belief.

\subsubsection{Belief Threshold of Delusion}

To systematically identify delusion, we define a belief threshold that distinguishes high-belief erroneous outputs from regular hallucinations. We set this threshold empirically by analyzing the belief distribution of correctly answered questions on a given dataset. Specifically, \textit{the belief threshold is determined as the mean belief level of all correctly answered questions}. When the model’s belief in an incorrect response surpasses this threshold, it is considered anomalously high and is classified as delusion. Intuitively, if an incorrect answer is assigned a belief level exceeding the average confidence of correct answers, it suggests an abnormal conviction in that error.

\begin{table*}[t]
    \footnotesize
    \centering
    \renewcommand{\arraystretch}{0.8}
    \begin{adjustbox}{width=1.0\textwidth}
    \begin{tabular}{cclcccccccccc}
        \toprule
        \multirow{5}{*}{\textbf{Model}} & \multirow{5}{*}{\textbf{Category}} & \multirow{5}{*}{\textbf{Method}} & \multicolumn{5}{c}{\textbf{TriviaQA}} & \multicolumn{5}{c}{\textbf{NQ}} \\
        \cmidrule(rl){4-8} \cmidrule(rl){9-13}
 & & & \multirow{3}{*}{\textbf{Acc.}} & \multirow{3}{*}{\textbf{ER}} & \multicolumn{3}{c}{\textbf{Delusion Ratio}} & \multirow{3}{*}{\textbf{Acc.}} & \multirow{3}{*}{\textbf{ER}} & \multicolumn{3}{c}{\textbf{Delusion Ratio}}\\
        \cmidrule(rl){6-8} \cmidrule(rl){11-13}
 & & & & & $\frac{\text{\#Delu}_\text{norm}}{\text{\#Error}}$ & $\frac{\text{\#Delu}}{\text{\#Error}}$ & $\frac{\text{\#Delu}}{\text{\#Total}}$ & & & $\frac{\text{\#Delu}_\text{norm}}{\text{\#Error}}$ & $\frac{\text{\#Delu}}{\text{\#Error}}$ & $\frac{\text{\#Delu}}{\text{\#Total}}$ \\
        \midrule
        \multirow{11}{*}{\textit{Llama-3.1-8B-Instruct}} & \multirow{5}{*}{\makecell[c]{Single \\Belief}}
 & logits & \multirow{5}{*}{69.3} & \multirow{5}{*}{28.8} & 22.0 & 25.5 &  7.3  & \multirow{5}{*}{51.7} & \multirow{5}{*}{45.3} & 31.0 & 28.7 & 13.0 \\
 & & consistency & & & 20.1 & 20.1 & 5.8 & & & 31.0 & 24.5 & 11.1 \\
 & & P(True) & & & 21.8 & 35.3 & 10.2 & & & 37.6 & 55.1 & 24.9 \\
 & & verb. 1S top-1 & & & 48.1 & 62.0 & 17.8 & & & 30.4 & 65.9 & 29.8 \\
 & & verb. 2S top-1 & & & 46.9 & 69.1 & 19.9 & & & 34.5 & 70.8 & 32.0 \\
        \cmidrule{2-13}
 & \multirow{4}{*}{\makecell[c]{Ensemble \\ Beliefs}}
 & P(True) \& consist. & \multirow{4}{*}{69.3} & \multirow{4}{*}{28.8} & 14.8 & 17.8 & 5.1 & \multirow{5}{*}{51.7} & \multirow{5}{*}{45.3} & 28.3 & 26.9 & 12.2 \\
 & & P(True) \& logits & & & 16.2 & 31.3 & 9.0 & & & 29.4 & 51.8 & 23.4 \\
 & & consist. \& logits & & & 15.9 & 20.1 & 5.8 & & & 26.7 & 24.4 & 11.1 \\
 & & P(True) \& consist. \& logits & & & 13.4 & 16.7 & 4.8 & & & 25.8 & 26.3 & 11.9 \\
        \midrule
        \multirow{11}{*}{\textit{Qwen2.5-7B-Instruct}} & \multirow{5}{*}{\makecell[c]{Single \\Belief}}
 & logits & \multirow{5}{*}{59.9} & \multirow{5}{*}{39.8} & 22.8 & 20.8 & 8.3 & \multirow{5}{*}{36.6} & \multirow{5}{*}{61.8} & 31.1 & 27.6 & 17.1 \\
 & & consistency & & & 32.1 & 32.1 & 12.8 & & & 31.5 & 31.5 & 19.4 \\
 & & P(True) & & & 19.2 & 48.6 & 19.4 & & & 34.0 & 50.2 & 31.0 \\
 & & verb. 1S top-1 & & & 77.8 & 77.8 & 31.0 & & & 65.6 & 65.6 & 40.5 \\
 & & verb. 2S top-1 & & & 77.0 & 76.7 & 30.6 & & & 66.0 & 66.0 & 40.7 \\
        \cmidrule{2-13}
 & \multirow{4}{*}{\makecell[c]{Ensemble \\ Beliefs}}
 & P(True) \& consist. & \multirow{4}{*}{59.9} & \multirow{4}{*}{39.8} & 13.9 & 24.9 & 9.9 & \multirow{5}{*}{36.6} & \multirow{5}{*}{61.8} & 25.1 & 30.4 & 18.8 \\
 & & P(True) \& logits & & & 15.4 & 45.7 & 18.2 & & & 25.6 & 48.9 & 30.2 \\
 & & consist. \& logits & & & 17.0 & 25.6 & 10.2 & & & 25.3 & 30.0 & 18.6 \\
 & & P(True) \& consist. \& logits & & & 12.6 & 22.7 & 9.0 & & & 22.5 & 29.8 & 18.4 \\
        \midrule
        \multirow{11}{*}{\textit{Mistral-7B-Instruct-v0.1}} & \multirow{5}{*}{\makecell[c]{Single \\Belief}}
 & logits & \multirow{5}{*}{53.6} & \multirow{5}{*}{46.1} & 27.0 & 31.7 & 14.6 & \multirow{5}{*}{30.5} & \multirow{5}{*}{68.1} & 28.5 & 24.1 & 16.4 \\
 & & consistency & & & 24.7 & 17.1 & 7.9 & & & 25.1 & 16.3 & 11.1 \\
 & & P(True) & & & 21.3 & 42.5 & 19.6 & & & 32.6 & 53.2 & 36.2 \\
 & & verb. 1S top-1 & & & 90.9 & 91.8 & 42.3 & & & 79.1 & 86.1 & 58.6 \\
 & & verb. 2S top-1 & & & 92.0 & 92.0 & 42.4 & & & 81.7 & 81.7 & 55.7 \\
        \cmidrule{2-13}
 & \multirow{4}{*}{\makecell[c]{Ensemble \\ Beliefs}}
 & P(True) \& consist. & \multirow{4}{*}{53.6} & \multirow{4}{*}{46.1} & 15.2 & 18.5 & 8.5 & \multirow{5}{*}{30.5} & \multirow{5}{*}{68.1} & 24.5 & 19.6 & 13.4 \\
 & & P(True) \& logits & & & 17.5 & 27.5 & 12.7 & & & 24.1 & 28.3 & 19.3 \\
 & & consist. \& logits & & & 20.1 & 19.4 & 8.9 & & & 24.1 & 18.4 & 12.5 \\
 & & P(True) \& consist. \& logits & & & 15.6 & 18.0 & 8.3 & & & 22.3 & 18.2 & 12.4 \\
        \midrule
        \multirow{11}{*}{\textit{Llama-3.3-70B-Instruct}} & \multirow{5}{*}{\makecell[c]{Single \\Belief}}
 & logits & \multirow{5}{*}{82.3} & \multirow{5}{*}{17.4} & 27.9 & 32.2 & 5.6 & \multirow{5}{*}{60.1} & \multirow{5}{*}{39.1} & 34.2 & 38.2 & 14.9 \\
 & & consistency & & & 53.2 & 53.2 & 9.2 & & & 56.2 & 56.2 & 22.0 \\
 & & P(True) & & & 31.3 & 77.8 & 13.5 & & & 35.9 & 84.2 & 32.9 \\
 & & verb. 1S top-1 & & & 38.8 & 38.8 & 6.7 & & & 28.6 & 67.8 & 26.5 \\
 & & verb. 2S top-1 & & & 54.6 & 67.9 & 11.8 & & & 47.5 & 70.6 & 27.6 \\
        \cmidrule{2-13}
 & \multirow{4}{*}{\makecell[c]{Ensemble \\ Beliefs}}
 & P(True) \& consist. & \multirow{4}{*}{82.3} & \multirow{4}{*}{17.4} & 25.9 & 44.8 & 7.8 & \multirow{5}{*}{60.1} & \multirow{5}{*}{39.1} & 33.6 & 48.8 & 19.1 \\
 & & P(True) \& logits & & & 24.0 & 64.0 & 11.1 & & & 31.2 & 73.6 & 28.8 \\
 & & consist. \& logits & & & 26.1 & 40.1 & 7.0 & & & 32.0 & 53.1 & 20.7 \\
 & & P(True) \& consist. \& logits & & & 23.3 & 44.4 & 7.7 & & & 29.3 & 49.8 & 19.4 \\
        \midrule
        \multirow{11}{*}{\textit{Qwen2.5-72B-Instruct}} & \multirow{5}{*}{\makecell[c]{Single \\Belief}}
 & logits & \multirow{5}{*}{75.2} & \multirow{5}{*}{24.7} & 22.7 & 23.4 & 5.8 & \multirow{5}{*}{50.6} & \multirow{5}{*}{49.2} & 31.4 & 31.7 & 15.6 \\
 & & consistency & & & 41.9 & 41.9 & 10.4 & & & 34.9 & 21.7 & 10.7 \\
 & & P(True) & & & 35.7 & 71.9 & 17.8 & & & 41.8 & 82.8 & 40.7 \\
 & & verb. 1S top-1 & & & 29.8 & 29.8 & 7.4 & & & 30.8 & 30.8 & 15.2 \\
 & & verb. 2S top-1 & & & 28.1 & 28.1 & 7.0 & & & 29.4 & 29.4 & 14.5 \\
        \cmidrule{2-13}
 & \multirow{4}{*}{\makecell[c]{Ensemble \\ Beliefs}}
 & P(True) \& consist. & \multirow{4}{*}{75.2} & \multirow{4}{*}{24.7} & 24.0 & 33.5 & 8.3 & \multirow{5}{*}{50.6} & \multirow{5}{*}{49.2} & 27.8 & 31.3 & 15.4 \\
 & & P(True) \& logits & & & 18.8 & 47.1 & 11.7 & & & 29.1 & 70.0 & 34.4 \\
 & & consist. \& logits & & & 20.0 & 33.9 & 8.4 & & & 28.3 & 23.7 & 11.7 \\
 & & P(True) \& consist. \& logits & & & 17.0 & 33.9 & 8.4 & & & 26.9 & 29.0 & 14.3 \\
        \bottomrule
    
    \end{tabular}
    \end{adjustbox}
    \vspace{-0.2cm}
    \caption{Delusion ratio based on different belief estimation strategies and their combinations.
    }
    
    \label{table:delusion-ratio-merged}
\end{table*}

\subsection{Delusion vs. Hallucination}
\label{sec:delusion_comparison}

Both delusion and hallucination in LLMs are fundamental errors in the model’s output, representing instances where the model fails to align its responses with factual accuracy.  A key distinction between hallucination and delusion lies in the confidence level of the model: \textbf{hallucinations can occur with low or moderate confidence, whereas delusions are identified by the model’s unwavering high confidence in its incorrect outputs.} Besides, recent psychiatric research~\cite{mourguescodern2024emergencedynamicsdelusionshallucinations} has shown that the re-emergence of delusions after remission is more common than hallucinations, which more often resolve first. This pattern mirrors our findings: we observe that delusions in models exhibit lower honesty and are harder to reject, with the model more likely to persist in its delusional beliefs during reflection (see ~\Cref{sec:comparison_delusion}).

\section{Empirical Study of Delusion}

This section presents an empirical study on delusion in LLMs. We investigate the distribution of delusions across datasets and models, with different belief estimation methods (\Cref{sec:distribution_delusion}) and belief ensemble techniques (\Cref{sec:belief_ensemble}). Furthermore, we discuss distinctions between delusions and hallucinations by exploring the challenges LLMs face in rejecting or reflecting upon delusions (\Cref{sec:comparison_delusion}).

\noindent\textbf{Experimental Setup.} We conduct experiments on two knowledge-based question-answering datasets: TriviaQA~\cite{joshi2017triviaqa} and Natural Questions (NQ, ~\citealp{kwiatkowski2019natural}). We use three well-known open-source model families:  \texttt{Qwen- 2.5-Instruct}\cite{qwen2.5} (1.5 / 3 / 7 / 14 / 30 / 70B), \texttt{Llama-3.1-8b-Instruct}\cite{dubey2024llama}, \texttt{Llama-3.3-70b-Instruct}, and \texttt{Mistral-7B-Instruct-v0.1}\cite{jiang2023mistral}. The specific details regarding the inference prompts, parameters, and fine-tuning procedures are provided in Appendix~\ref{sec:empirical_study_setup}.

\noindent\textbf{Evaluation Metrics.} We primarily evaluate the overall performance using accuracy and error rate (ER). Model responses are categorized into three types: correct, incorrect, and rejected, with rejections being relatively rare. Incorrect responses are further classified into delusions and hallucinations based on a specific belief score. We use the proportion of delusions within the entire dataset or among all incorrect responses as delusion metrics. For different belief calculation methods, we use the same model response obtained through greedy search.

\subsection{Distribution of Delusion}
\label{sec:distribution_delusion}
As illustrated in Table~\ref{table:delusion-ratio-merged}, empirical findings reveal consistently high delusion ratios across diverse model architectures and evaluation strategies, suggesting this phenomenon constitutes a pervasive challenge transcending model configurations. For instance, in TriviaQA under non-ensemble belief conditions, the \texttt{Qwen2.5} (7B) model exhibits an overall delusion rate ranging from 8.3\%–31.0\%, with delusions accounting for 20.8\%–77.8\% of all erroneous responses. Notably, scaling to the \texttt{Qwen2.5} (72B) architecture reduces overall delusion rates to 5.8\%–17.8\%, demonstrating that enhanced model capacity improves response fidelity. However, delusions persist in 23.4\%–79.1\% of erroneous outputs, indicating that architectural scaling primarily elevates the general hallucination phenomenon while showing limited efficacy in mitigating delusional tendencies in erroneous responses.

Among all the evaluation metrics, verbal-based methods exhibit significantly higher rates of delusion, highlighting the inherent challenges in relying on verbal responses for belief estimation. These methods show a more pronounced discrepancy between expected and actual outcomes, suggesting that alternative approaches may be needed to address these issues more effectively.



\subsection{Can Belief Ensemble Eliminate Delusions?}

\label{sec:belief_ensemble}

To assess whether combining different belief estimation methods can mitigate delusions, we ensemble three of the most effective belief estimation techniques: P(True), consistency, and logits. Our findings indicate that ensemble methods do help reduce delusions. For instance, in the TriviaQA dataset, delusion ratios in errors dropped from 71.9\%-33.5\% in \texttt{Qwen2.5} (72B). However, despite these improvements, this suggests that while ensemble techniques provide some benefit, belief estimation itself is not the primary cause of delusions. Other factors, possibly inherent to the model's architecture or the nature of the task, likely contribute significantly to the high delusion rates observed.

\subsection{Differences between Delusions and Hallucinations}
\label{sec:comparison_delusion}
As discussed in \Cref{sec:delusion_comparison} and in the psychiatric literature~\cite{mourguescodern2024emergencedynamicsdelusionshallucinations}, delusions are more difficult to address and are more prone to re-emergence than hallucinations. Therefore, here we empirically validate the distinction between delusions and hallucinations. The specific experimental setup can be found in the Appendix~\ref{sec:delu_vs_hallu}.

\subsubsection{LLMs Show Less Honesty with Delusions}

LLMs exhibit less honesty when dealing with delusions. We evaluate their willingness to refuse to answer unknown questions using prompts representing different honesty levels (see Appendix~\ref{sec:Prompt_Reject}), as shown in Figure~\ref{fig:by_prompt}. The results indicate that while the overall error rate changes with different prompts, the delusion refusal rate consistently remains lower than the hallucination across all settings. This suggests that LLMs are more inclined to reject normal hallucinations but are less willing to refuse delusional questions. Furthermore, as depicted in Figure~\ref{fig:by_size}, although larger models show increased accuracy and decreased error rate, the delusion refusal rate still consistently surpasses the hallucination refusal rate. These findings highlight that LLMs tend to have a stronger internal belief in delusions, making them less likely to reject delusional content compared to hallucinations.

\begin{figure}[h!]
    \centering
    \includegraphics[width=\linewidth,trim=0 0 0 0,clip]{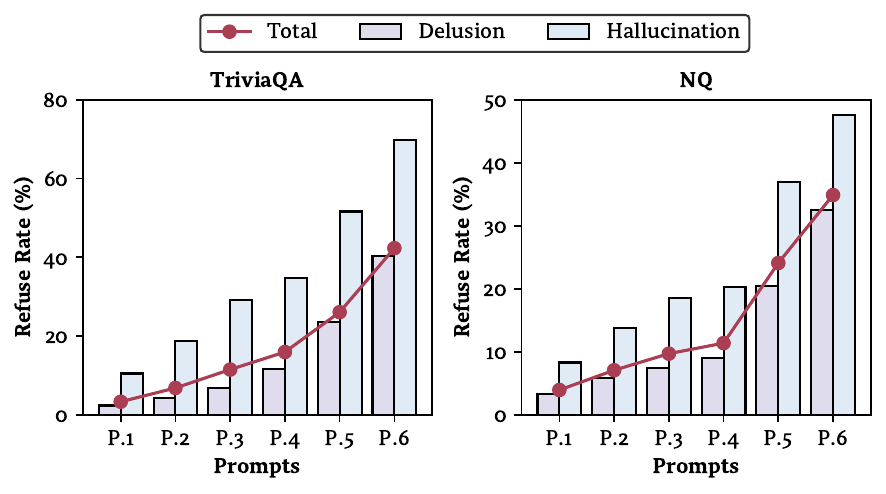}
    \vspace{-0.4cm}
    \caption{Refuse ratio comparison with prompts of different honesty levels. Model: Llama3.1-8b-instruct. See Appendix~\ref{sec:complete_result_by_prompt} for complete results of three models.}
    \label{fig:by_prompt}
\end{figure}

\begin{figure}[h!]
    \centering
    \includegraphics[width=\linewidth,trim=0 0 0 0,clip]{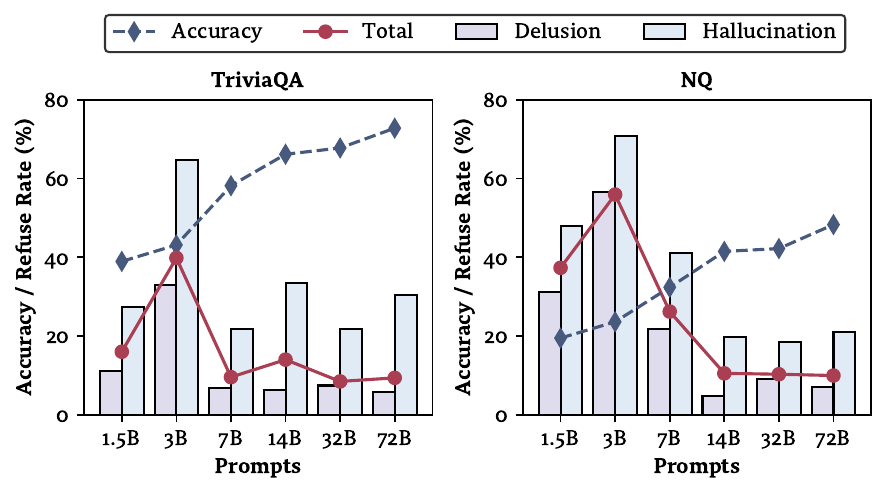}
    \vspace{-0.4cm}
    \caption{Refuse rate comparison with different model sizes. Models: Qwen2.5 family.}
    \label{fig:by_size}
\end{figure}



\subsubsection{LLMs Struggle More to Reject Delusions}

\begin{figure}[h!]
    \centering
    \includegraphics[width=\linewidth,trim=0 0 0 0,clip]{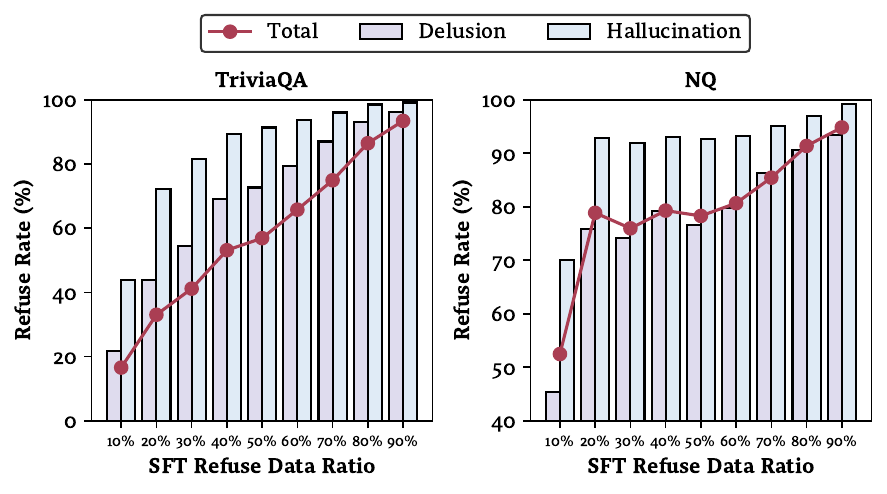}
    \vspace{-0.4cm}
    \caption{Refuse rate comparison by different SFT refuse data ratio. Model: Llama3.1-8b-instruct. Refer to Appendix~\ref{sec:complete_result_by_ratio} for complete results of three models.} 
    \label{fig:by_ratio}
\end{figure}

We further modify the incorrect outputs by replacing them with "I don't know" and fine-tune the models to learn to reject unknown questions. The training set is constructed by adjusting the proportion of rejection data, which includes both delusions and hallucinations, to create a balanced training set. As shown in Figure~\ref{fig:by_ratio}, the results demonstrate that even after training, the models remain more inclined to reject normal hallucinations while retaining a higher proportion of delusions. This suggests that while the models improve their ability to reject certain types of erroneous outputs, they continue to struggle with rejecting delusional content, maintaining a stronger internal belief in delusions compared to hallucinations.

\subsubsection{Delusions Are Harder to Reflect Upon than Hallucinations}

\begin{figure}[t!]
    \centering
    \includegraphics[width=\linewidth,trim=0 0 0 0,clip]{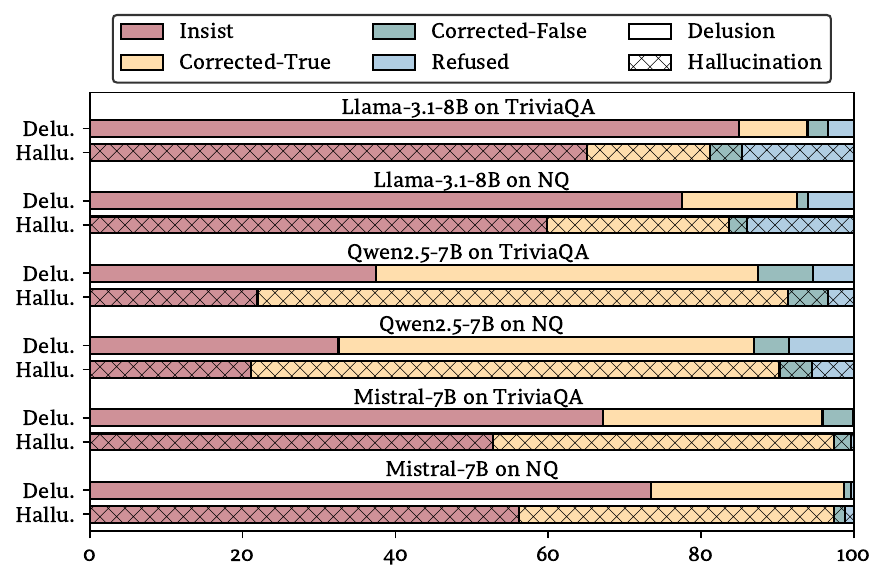}
    \vspace{-0.4cm}
    \caption{Distribution of reflection outcomes on delusions and hallucinations. LLMs tend to insist more on delusions than hallucinations.} 
    \label{fig:reflect}
\end{figure}

We prompt models to reflect on their previous answers and assess whether they are willing to change their initial responses. The results, as shown in Figure~\ref{fig:reflect}, indicate that models are significantly more likely to insist on their delusional responses than they are on non-delusional hallucinations. This behavior suggests that LLMs exhibit a stronger internalized belief in delusions, as they are more resistant to revising their answers when confronted with delusional content. In contrast, when dealing with hallucinations, models show a higher willingness to reconsider their responses, reflecting a less entrenched belief in these types of errors.

\section{Causes and Influence Factors}
In this section, we primarily investigate the causes of delusion formation and the factors that influence changes in delusion. Our research focuses on two main aspects: data and training. The experiments are mainly conducted on the ALCUNA~\cite{yin-etal-2023-alcuna} dataset, where all the questions are related to fictional entities. This allows us to easily introduce noise and make other modifications without worrying about the impact of the model's pre-existing knowledge on the results. The specific training setup can be found in the Appendix~\ref{sec:dynamics_setup}.
\label{sec:dynamics}

\subsection{The Effect of Data Noise on Delusions}

\begin{figure}
    \centering
    \includegraphics[width=\linewidth]{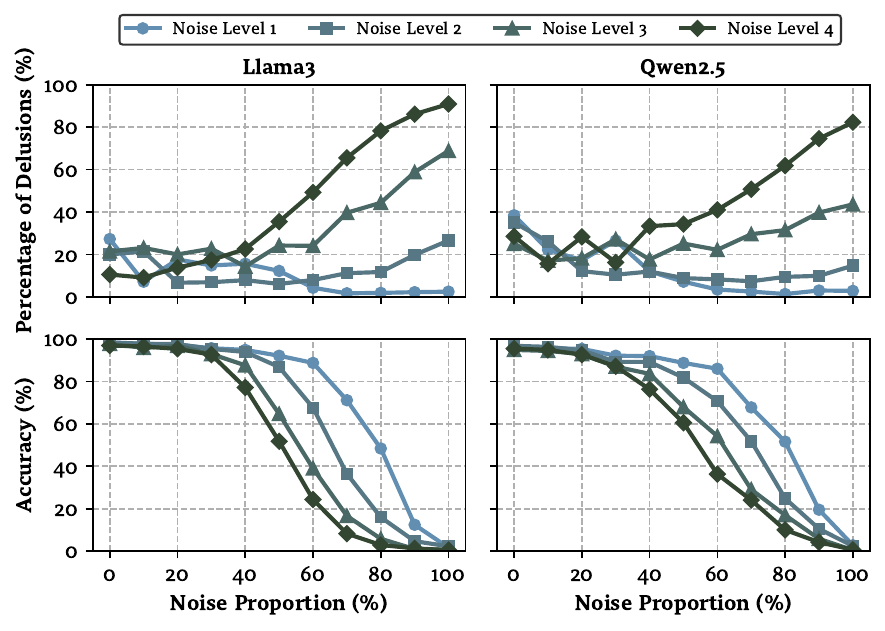}
    \vspace{-0.4cm}
    \caption{Comparison of delusion percentage and accuracy with different noise proportion and noise level.}
    \label{fig:noise proportion}
\end{figure}

\noindent\textbf{Noise Proportion.} The impact of data noise on delusion formation is analyzed by training models with different proportions of noisy and clean data. As shown in Figure~\ref{fig:noise proportion}, the x-axis represents the proportion of noisy data mixed with clean data during training, while the y-axis displays both the delusion rate and the model's prediction accuracy. The results indicate that as the proportion of noisy data increases, the delusion rate rises, while the accuracy of the model decreases.

\noindent\textbf{Noise intensity.} We investigate how noise intensity affects delusion formation. High noise intensity is characterized by a concentrated distribution of erroneous answers, whereas low noise intensity results in a more dispersed distribution. Our findings reveal that when the noisy data has a higher intensity (more concentrated errors), the delusion rate increases more significantly as the proportion of noisy data rises, leading to a faster decline in accuracy. In contrast, when the noise intensity is lower, the delusion rate increases more gradually, and accuracy decreases at a slower pace. Interestingly, under low noise intensity conditions, although the error rate decreases as more noisy data is added, the delusion rate also declines. This suggests that when errors are distributed more evenly, the model is less likely to form strong beliefs in any particular incorrect answer, leading to a higher rate of hallucinations but lower delusion occurrence.

\subsection{How Training Affects Delusions?}

Our analysis further reveals that even when the proportion of noisy data is set to zero, some experiments still exhibit a high delusion rate. This suggests that delusions may arise in the model's default state, even before the introduction of noise. Upon examining the training samples using the \texttt{paraphrase-MiniLM-L6-v2}~\cite{reimers-gurevych-2019-sentence} model for embedding and calculating cosine similarity, we found that certain questions in the training set had high similarity scores with one another. These questions, paired with answers that closely aligned with known delusion examples, created interference in the model’s learning process. On average, each delusion example had 27 other training data points with similar questions and identical answers, as shown in Figure~\ref{fig:similar_data}. We believe that this interference in the data may have disrupted the model’s training, increasing the difficulty of the training and learning process. We further validated our hypothesis through the following two experimental approaches:

\begin{table}[htbp]
    \centering
    \resizebox{\linewidth}{!}{
    \begin{tabular}{lcclll}
        \toprule
        \multirow{2.7}{*}{\textbf{Data Formation}}& \multirow{2.7}{*}{\textbf{Acc.}} & \multirow{2.7}{*}{\textbf{ER}} & \multicolumn{3}{c}{\textbf{Delusion Ratio}}  \\
        \cmidrule(rl){4-6}
         & & & $\frac{\text{\#Delu}_\text{norm}}{\text{\#Error}}$ & $\frac{\text{\#Delu}}{\text{\#Error}}$ & $\frac{\text{\#Delu}}{\text{\#Total}}$ \\
         \midrule
         Full Data & 93.2 & 6.8 & 3.9 & 9.9 & 0.7 \\
         \ \ \ w/o similar data & 96.2 & 3.8 & $\text{1.1}_{(\textcolor{red}{-71.3\%})}$ & $\text{3.8}_{(\textcolor{red}{-62.2\%})}$  & $\text{0.1}_{(\textcolor{red}{-78.7\%})}$  \\
         \bottomrule
    \end{tabular}}
    \vspace{-0.2cm}
    \caption{Model Performance on data ablation study.}
    \label{tab:deduplicate}
\end{table}

\begin{figure}
    \centering
    \includegraphics[width=\linewidth]{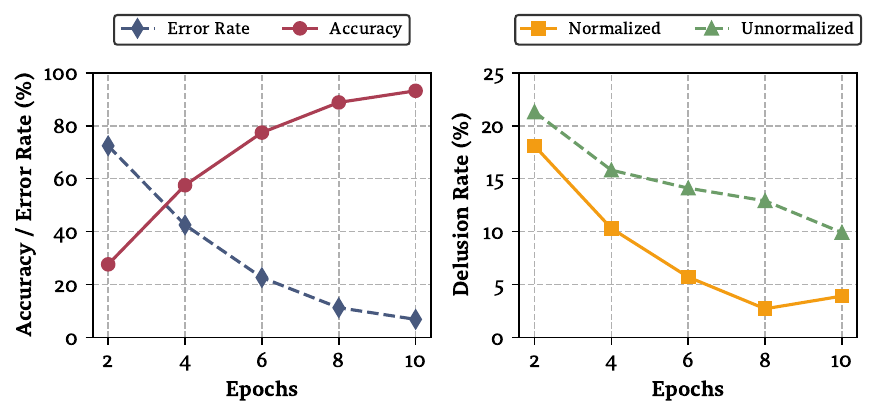}
    \vspace{-0.4cm}
    \caption{Trends of metrics across training epochs.}
    \label{fig:progress}
\end{figure}

\noindent \textbf{1) Reducing training interference.} when we refined the training set by removing question-answer pairs with high cosine similarity scores (greater than 0.9) to known delusions, the delusion rate significantly decreased as shown in Table~\ref{tab:deduplicate}. Retraining the model with this refined dataset demonstrated that dataset refinement plays a crucial role in reducing delusions. 
\noindent \textbf{2) Improving training sufficiency.} In Figure~\ref{fig:progress}, we observed that as training progressed, the model’s accuracy improved, and its error rate decreased, while the delusion rate (both normalized and unnormalized) showed a marked decline. These findings highlight that sufficient training, alongside careful management of dataset interference, is effective in reducing the occurrence of delusions during the model's training process.

\section{Mitigating Delusions via External Verification}
\label{sec:mitigating}
In this section, we explore how to mitigate delusions. Since delusions are more difficult to detect using uncertainty-based methods compared to hallucinations, we attempt to address this issue by introducing external knowledge for answer validation. Specifically, we explore two approaches in the following subsections: model debating and Retrieval-Augmented Generation (RAG). The detailed experimental setup can be found in Appendix~\ref{sec:external_verify_setup}.

\subsection{Mitigating Delusions through Multi-agent Debating}

\begin{table}[!ht]

      \small
    \centering
    \renewcommand{\arraystretch}{1.0}

    \begin{adjustbox}{width=\linewidth}
    \begin{tabular}{lccccll}
        \toprule
\multirow{3}{*}{\textbf{Methods}} & \multirow{3}{*}{\textbf{Acc.}} & \multirow{3}{*}{\textbf{ER}} & \multicolumn{3}{c}{\textbf{Delusion Ratio}}  & \multirow{3}{*}{\textbf{Hallu. Ratio}}\\
         \cmidrule(rl){4-6}
        &  &  & $\frac{\text{\#Delu}_\text{norm}}{\text{\#Error}}$ & $\frac{\text{\#Delu}}{\text{\#Error}}$ & $\frac{\text{\#Delu}}{\text{\#Total}}$ \\
        \midrule
        \rowcolor{gray!15} \multicolumn{7}{c}{\textit{llama-3.1-8b-instruct}} \\
        Original & 69.3 & 28.8 & 22.0 & 25.5 & 7.3 & 21.5\\
        + Vote 1/3 & 58.8 & 12.0  & 13.3 & 15.0 & $\text{4.3}_{(\textcolor{red}{-41.0\%})}$& $\text{7.6}_{(\textcolor{blue}{-64.5\%})}$\\
        + Vote 2/3 & 47.2 & 6.3  & 8.6 & 9.6 & $\text{2.8}_{(\textcolor{red}{-62.2\%})}$& $\text{3.5}_{(\textcolor{blue}{-83.7\%})}$\\
        + Vote 3/3 & 31.1 & 2.7  & 3.8 & 4.2 & $\text{1.2}_{(\textcolor{red}{-83.5\%})}$& $\text{1.5}_{(\textcolor{blue}{-93.2\%})}$\\
        \rowcolor{gray!15} \multicolumn{7}{c}{\textit{Qwen2.5-7B-Instruct}} \\
        Original & 59.9 & 39.8 & 22.8 & 20.8 & 8.3 & 31.5\\
        + Vote 1/3 & 46.5 & 11.5  & 9.5 & 8.8 & $\text{3.5}_{(\textcolor{red}{-57.5\%})}$& $\text{8.0}_{(\textcolor{blue}{-74.8\%})}$ \\
        + Vote 2/3 & 41.9 & 6.2  & 6.0 & 5.7 & $\text{2.3}_{(\textcolor{red}{-72.8\%})}$& $\text{4.0}_{(\textcolor{blue}{-87.3\%})}$\\
        + Vote 3/3 & 32.1 & 2.9  & 3.1 & 3.0 & $\text{1.2}_{(\textcolor{red}{-85.6\%})}$& $\text{1.7}_{(\textcolor{blue}{-94.5\%})}$\\
        \rowcolor{gray!15} \multicolumn{7}{c}{\textit{Mistral-7B-Instruct-v0.1}} \\
        Original & 53.6 & 46.1 & 27.0 & 31.7 & 14.6 & 31.5 \\
        + Vote 1/3 & 45.9 & 12.4  & 9.5 & 10.8 & $\text{5.0}_{(\textcolor{red}{-65.9\%})}$& $\text{7.5}_{(\textcolor{blue}{-76.3\%})}$ \\
        + Vote 2/3 & 40.6 & 6.1  & 4.9 & 5.5 & $\text{2.5}_{(\textcolor{red}{-82.6\%})}$ & $\text{3.5}_{(\textcolor{blue}{-88.8\%})}$ \\
        + Vote 3/3 & 30.0 & 2.6  & 2.4 & 2.8 & $\text{1.3}_{(\textcolor{red}{-91.3\%})}$& $\text{1.3}_{(\textcolor{blue}{-95.8\%})}$ \\
        \rowcolor{gray!15} \multicolumn{7}{c}{\textit{gpt3.5-turbo}} \\
        Original & 71.2 & 28.8 & 15.8 & 28.7 & 8.3 & 20.5\\
        + Vote 1/3 & 62.2 & 11.5  & 12.5 & 20.2 & $\text{5.8}_{(\textcolor{red}{-29.6\%})}$& $\text{5.7}_{(\textcolor{blue}{-72.3\%})}$\\
        + Vote 2/3 & 48.9 & 5.9  & 8.7 & 12.7 & $\text{3.7}_{(\textcolor{red}{-55.5\%})}$& $\text{2.2}_{(\textcolor{blue}{-89.2\%})}$\\
        + Vote 3/3 & 31.9 & 2.5  & 4.3 & 5.9 & $\text{1.7}_{(\textcolor{red}{-79.5\%})}$& $\text{0.8}_{(\textcolor{blue}{-95.9\%})}$\\
        \bottomrule
    \end{tabular}%
    \end{adjustbox}
 \caption{Delusion mitigation across different models and thresholds through multi-agent debating.}
   \label{table:vote}
\end{table}

We employed a multi-agent voting approach where each model’s output is validated by the other models. This voting approach helps filter out erroneous answers by leveraging the collective knowledge of multiple models. Experimental results, as shown in Table~\ref{table:vote}, demonstrate that multi-agent voting effectively mitigates delusions, reducing error rates and delusion ratios across all models. Notably, Mistral-7B achieved the largest reduction, from 14.6\% to 1.3\%. While delusions are generally harder to mitigate than hallucinations, a significant portion of delusions was still addressed, highlighting the effectiveness of multi-agent voting in reducing both delusions and hallucinations. These findings suggest that multi-agent voting is a robust method for mitigating delusions and improving model reliability.

\subsection{Mitigating Delusions through RAG}

\begin{table}[!ht]

    \small
    \centering
    \renewcommand{\arraystretch}{1.0}
    
    \begin{adjustbox}{width=\linewidth}
    \begin{tabular}{lccccll}
        \toprule
\multirow{3}{*}{\textbf{Methods}} & \multirow{3}{*}{\textbf{Acc.}} & \multirow{3}{*}{\textbf{ER}} & \multicolumn{3}{c}{\textbf{Delusion Ratio}} & \multirow{3}{*}{\textbf{Hallu. Ratio}}\\
         \cmidrule(rl){4-6}
        &  &  & $\frac{\text{\#Delu}_\text{norm}}{\text{\#Error}}$ & $\frac{\text{\#Delu}}{\text{\#Error}}$ & $\frac{\text{\#Delu}}{\text{\#Total}}$ \\
        \midrule
        \rowcolor{gray!15} \multicolumn{7}{c}{\textit{llama-3.1-8b-instruct}} \\
        original & 69.3 & 28.8 & 22.0 & 25.5 & 7.3 & 21.5 \\
        \ + RAG & 88.2 & 9.5  & 7.8 & 9.0 & $\text{2.6}_{(\textcolor{red}{-64.7\%})}$& $\text{6.9}_{(\textcolor{blue}{-67.9\%})}$ \\
        \rowcolor{gray!15} \multicolumn{7}{c}{\textit{Qwen2.5-7B-Instruct}} \\
        original & 59.9 & 39.8 & 22.8 & 20.8 & 8.3 & 31.6 \\
        \ + RAG & 87.6 & 12.4  & 7.1 & 6.7 & $\text{2.7}_{(\textcolor{red}{-67.8\%})}$& $\text{9.7}_{(\textcolor{blue}{-69.3\%})}$  \\
        \rowcolor{gray!15} \multicolumn{7}{c}{\textit{Mistral-7B-Instruct-v0.1}} \\
        original & 53.6 & 46.1 & 27.0 & 31.7 & 14.6 & 31.5\\
        \ + RAG & 81.7 & 18.3  & 7.7 & 8.9 & $\text{4.1}_{(\textcolor{red}{-71.8\%})}$& $\text{14.2}_{(\textcolor{blue}{-55.1\%})}$ \\
        \bottomrule
    \end{tabular}
    \end{adjustbox}
    \caption{Delusion mitigation through RAG.}
\label{table:rag}
\end{table}
We apply RAG to TriviaQA to mitigate delusions. For each question, models use 20 relevant passages retrieved from a knowledge base to enhance their responses. This allows the model to augment its generation process with external knowledge, helping to mitigate delusions by grounding answers in relevant information. The results, as shown in Table~\ref{table:rag}, demonstrate that this method significantly reduces the delusion rate across all models. This indicates that incorporating retrieval improves the factual accuracy of model-generated answers, mitigating the occurrence of delusions. Moreover, the reduction in delusions is comparable to the reduction in hallucinations, highlighting that RAG is an effective method for addressing both delusions and hallucinations.

\section{Conclusion}
This paper introduces the concept of LLM delusion, a more insidious and persistent phenomenon compared to traditional hallucinations in LLMs. We demonstrate that delusions, characterized by high belief in factually incorrect responses, pose a unique challenge due to their low uncertainty and resistance to detection. Our empirical analysis across multiple LLM families reveals the widespread presence of delusions, highlighting the need for targeted mitigation strategies. We show that while self-reflection and fine-tuning methods have limited success in reducing delusions, external verification approaches such as RAG and multi-agent debate systems offer potential solutions. These findings underscore the importance of robust model validation and confidence calibration to improve the reliability and trustworthiness of LLMs in real-world applications.

\section*{Limitations}

This study introduces the concept of delusion and investigates it in mainstream open-source large models. However, some uncertainty methods face challenges in accessing all necessary information (such as specific token logits) on closed-source models, and this paper does not explore delusion in such models. Additionally, the experiments are primarily focused on knowledge-based question-answering datasets, leaving delusion in other tasks underexplored. Moreover, while delusion has been extensively studied in the field of mental health, this paper could not integrate these studies due to space limitations.

\bibliography{custom,anthology}
\newpage
\appendix

\section{Uncertainty Estimation Methods}
\label{sec:uncertainty}
Below is a detailed summary of the uncertainty estimation methods used in this study:

\begin{enumerate}
    \item \textbf{Raw Logits}: This method uses the logit values to estimate confidence, taking the exponential of the average log probability of tokens. It’s equivalent to the reciprocal of perplexity, representing how "certain" the model is about its prediction.
    \item \textbf{Agreement (Consistency-based)}: Confidence is calculated by the percentage of answers in a set that agree with the most-voted answer, reflecting the model's consistency in its outputs.
    \item \textbf{P(True)}: The model evaluates the truthfulness of its response, with the confidence being the normalized probability assigned to the ‘True’ token.
    \item \textbf{Verb. 1S Top-k}: The model generates k possible answers and their probabilities in one step. The top answer and its probability represent the model’s confidence.
    \item \textbf{Verb. 2S Top-k}: This two-stage method has the model first generate possible answers and then assign probabilities to them in a second round. The final confidence is based on these probabilities.
\end{enumerate} 

These methods offer a range of strategies to quantify the model's belief in its outputs, with different approaches to capturing certainty and uncertainty. Prompts used in these methods are shown in Appendix~\ref{sec:Prompt_Belief}.

\section{Experimental Setup}
\subsection{Empirical Study of Delusion}
\label{sec:empirical_study_setup}
\subsubsection{Distribution of Delusion}
We employed the five confidence estimation methods mentioned in Appendix~\ref{sec:uncertainty}, with the exception of the consistency method. For the other methods, inference was performed without sampling, and the max\_tokens was set to 128. The consistency method used a sampling temperature of 0.7, top\_p of 0.95, and top\_k of 40.

\subsubsection{Can Belief Ensemble Eliminate Delusions?}
In the Belief Ensemble experiments, we averaged the outputs from different belief estimation methods and subsequently used the new belief scores to calculate the delusion rate. 

\subsection{Delusion vs. Hallucination}
\label{sec:delu_vs_hallu}
\subsubsection{LLMs Show Less Honesty with Delusions}
In the experiment designed to assess the models' willingness to reject unknown questions, we guided the models with different prompt strategies. The detailed prompt formulations can be found in the Appendix~\ref{sec:Prompt_Reject}. To examine the impact of various prompts on the delusion rejection rate, we utilized six different prompts. These prompts were designed to guide the models in rejecting questions that they could not confidently answer. In the investigation of the effect of model size on rejection rates, we employed the same prompt across all models, ensuring consistency in the rejection behavior analysis. This experiment was aimed at investigating how well the models could distinguish between answering and rejecting questions that fall outside their knowledge scope.

\subsubsection{LLMs Struggle More to Reject Delusions}
For the supervised fine-tuning (SFT) experiments focused on improving the models' ability to reject questions, we extracted 10,000 data points from the TriviaQA training set. This dataset contained a specific mix of correct and incorrect answers, ranging from a 1:9 to a 9:1 ratio. Incorrect answers were labeled with "I don’t know," while correct answers were assigned their corresponding labels. The models were fine-tuned using an SFT approach with 2 epochs and a learning rate of 
$1e^{-5}$.

\subsubsection{Delusions Are Harder to Reflect Upon than Hallucinations}
In the model reflection experiment, we prompted the models to reflect on their initial answers by including the first-round answer in the prompt. The models were then guided to reconsider the correctness of their answers. The prompt details are provided in the Appendix~\ref{sec:Prompt_reflection}.

\subsection{Formation and Dynamics of Delusion}
\label{sec:dynamics_setup}
\subsubsection{The Effect of Data Noise on Delusions}
In this study, we augment the ALCUNA dataset by introducing noise. Specifically, for each correct question-answer pair, we generate 20 perturbed incorrect answers. This augmentation process simulates the influence of erroneous information on the model's ability to distinguish between correct and incorrect answers.

For the fine-tuning process, we select a subset of 30,271 numeric-answer questions from the ALCUNA dataset, which includes numerical answers and multiple-choice question options. The dataset is divided into training (27,280 samples) and testing sets (test sets not used during training). The training set is further divided into 11 noise proportions (ranging from 0\% to 100\%), with each noise proportion corresponding to four noise intensity levels (NoiseLevel). Each noise level contains 620 samples, and for each sample, 20 variations are generated via supervised fine-tuning (SFT). The noise is generated by randomly modifying the correct answer through the addition, deletion, or modification of characters. For each correct answer, perturbations are made by applying these changes to the text, which introduces variability into the data. The noise levels, defined as NoiseLevel, control the degree of consistency among the perturbed answers. The noise introduced at each level varies as follows: for NoiseLevel 4, 100\% of the noise consists of answers that are identical to one another, generated by modifying the same data point. For NoiseLevel 3, 75\% of the noisy answers share the same modified version of the data, while the remaining 25\% are different. In NoiseLevel 2, 50\% of the noisy answers are consistent, and in NoiseLevel 1  `, 25\% of the noisy answers are identical, with the remaining answers being randomly altered from the standard answer. This ensures that the noise is spread across a range of data points at different intensities.

The SFT process is conducted with 5 epochs and a learning rate of $3 \times 10^{-6}$. During the training phase, we sample from these augmented pairs to simulate different levels of noise and evaluate its impact on delusion formation. By varying the noise proportion and intensity, we aim to explore how the distribution of noisy data and the fine-tuning process affect the model's ability to form delusions.

\subsubsection{How Training Affects Delusions}
In investigating the impact of the training process on delusion formation, we constructed supervised fine-tuning (SFT) training data using 30,271 samples from ALCUNA with entirely correct labels. The model was trained for 10 epochs with a learning rate of $1 \times 10^{-5}$.

To assess the effect of reducing training interference, we first used the paraphrase-MiniLM-L6-v2 model\footnote{ https://huggingface.co/sentence-transformers/paraphrase-MiniLM-L6-v2} to generate embeddings for all samples. We then computed the cosine similarity between each sample’s embedding and the embeddings of the 206 delusion examples identified within the 30,271 samples. If a sample shared the same answer as any delusion example or had a cosine similarity score greater than 0.9 with any delusion sample, it was removed from the dataset. After this filtering process, 23,507 samples remained, and these were used for training with their correct labels for 10 epochs, with a learning rate of $1 \times 10^{-5}$.

\begin{figure}
    \centering
    \includegraphics[width=\linewidth]{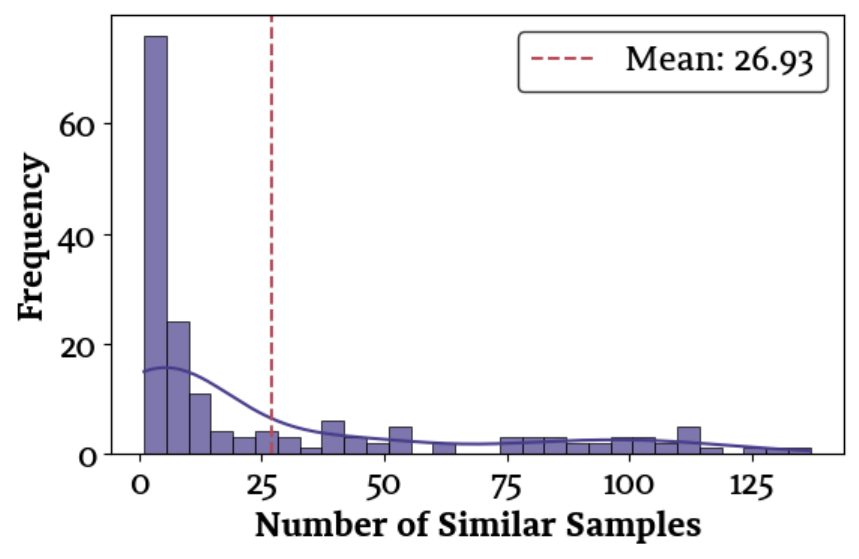}
    \caption{Distribution of the number of similar samples for each delusion sample.}
    \label{fig:similar_data}
\end{figure}

\subsection{Mitigating Delusions via External Verification}
\label{sec:external_verify_setup}
\subsubsection{Mitigating Delusions through Multi-agent Debating}

For each question in TriviaQA, one model was designated as the "target" model, and the other models were used as "verifiers". The target model’s answer was validated by comparing it against the responses from the verifier models. To determine whether the answer should be accepted as correct, a predefined threshold (ranging from 1 to 3) was applied. This threshold indicated the minimum number of verifier models that needed to agree with the target model’s response for it to be considered accurate. If the number of matching answers from the verifiers fell below the threshold, the target model's response was classified as a delusion and discarded. 

\subsubsection{Mitigating Delusions through RAG}
For each question, 20 relevant passages are retrieved from a knowledge base to assist in generating the answer. The passages used in this experiment are those extracted for TriviaQA in the InstructRAG~\cite{wei2024instructrag} framework. The prompt used in this experiment can be found in Appendix~\ref{sec:Prompt_rag}.

\section{Complete Experiment Results}

\subsection{Refuse Rate Comparison with Prompts of Different Honesty Levels}
\label{sec:complete_result_by_prompt}
The complete results are shown in Table~\ref{table:by_prompts}.

\begin{table*}[t]
\small
\centering
\setlength\tabcolsep{5pt}
\renewcommand{\arraystretch}{0.8}
\begin{tabular}{lccccccc}
\toprule
\multirow{2.5}{*}{\textbf{Model}} & \multirow{2.5}{*}{\textbf{Prompt Strategy}} & \multicolumn{3}{c}{\textbf{TriviaQA}} & \multicolumn{3}{c}{\textbf{NQ}} \\
\cmidrule(rl){3-5} \cmidrule(rl){6-8}
& & \textbf{Delu.} & \textbf{Hallu.} & \textbf{Total} & \textbf{Delu.} & \textbf{Hallu.} & \textbf{Total} \\
\midrule
\multirow{6}{*}{\textit{Llama-3.1-8B-Instruct}}
& \texttt{helpful\_can\_refuse} & 4.3 & 18.8 & 6.8 & 5.8 & 13.8 & 7.1 \\
& \texttt{helpful\_less\_refuse} & 2.4 & 10.5 & 3.3 & 3.3 & 8.3 & 4.0 \\
& \texttt{helpful\_more\_refuse} & 6.9 & 29.2 & 11.5 & 7.5 & 18.6 & 9.7 \\
& \texttt{helpful\_medium\_refuse} & 23.5 & 51.6 & 26.1 & 20.4 & 37.0 & 24.1 \\
& \texttt{helpful\_most\_refuse} & 40.3 & 69.7 & 42.3 & 32.5 & 47.6 & 34.9 \\
& \texttt{helpful\_high\_refuse} & 11.7 & 34.9 & 16.0 & 9.0 & 20.3 & 11.4 \\
\midrule
\multirow{6}{*}{\textit{Qwen2.5-7B-Instruct}}
& \texttt{helpful\_can\_refuse} & 7.6 & 24.1 & 10.6 & 15.0 & 32.2 & 20.1 \\
& \texttt{helpful\_less\_refuse} & 2.2 & 11.3 & 4.6 & 8.3 & 18.6 & 11.1 \\
& \texttt{helpful\_more\_refuse} & 7.9 & 23.0 & 10.5 & 23.8 & 43.3 & 28.4 \\
& \texttt{helpful\_medium\_refuse} & 6.8 & 21.8 & 9.6 & 21.9 & 41.1 & 26.2 \\
& \texttt{helpful\_most\_refuse} & 24.1 & 48.2 & 24.7 & 49.9 & 70.6 & 52.0 \\
& \texttt{helpful\_high\_refuse} & 23.4 & 49.7 & 25.4 & 51.7 & 73.1 & 54.5 \\
\midrule
\multirow{6}{*}{\textit{Mistral-7B-Instruct-v0.1}}
& \texttt{helpful\_can\_refuse} & 1.6 & 14.3 & 6.1 & 6.1 & 21.4 & 14.1 \\
& \texttt{helpful\_less\_refuse} & 0.7 & 8.1 & 3.3 & 3.0 & 14.1 & 8.6 \\
& \texttt{helpful\_more\_refuse} & 28.8 & 52.4 & 32.0 & 43.4 & 68.6 & 54.9 \\
& \texttt{helpful\_medium\_refuse} & 8.4 & 28.3 & 14.1 & 18.8 & 43.0 & 31.1 \\
& \texttt{helpful\_high\_refuse} & 25.7 & 52.4 & 30.8 & 39.3 & 68.3 & 54.2 \\
& \texttt{helpful\_most\_refuse} & 13.4 & 36.2 & 19.4 & 24.2 & 50.1 & 36.8 \\
\bottomrule
\end{tabular}
\caption{Complete results of refuse rate comparison with prompts of different honesty levels.}
\label{table:by_prompts}
\end{table*}

\subsection{Refuse Rate Comparison with Different SFT Refuse Data Ratio}
\label{sec:complete_result_by_ratio}
The complete results are shown in Table~\ref{table:by_ratio}

\begin{table*}[t]
\small
\centering
\setlength\tabcolsep{5pt}
\renewcommand{\arraystretch}{0.8}
\begin{tabular}{lccccccc}
\toprule
\multirow{2.5}{*}{\textbf{Model}} & \multirow{2.5}{*}{\makecell[c]{\textbf{SFT Refuse}\\\textbf{Data Ratio}}} & \multicolumn{3}{c}{\textbf{TriviaQA}} & \multicolumn{3}{c}{\textbf{NQ}} \\
\cmidrule(rl){3-5} \cmidrule(rl){6-8}
& & \textbf{Delu.} & \textbf{Hallu.} & \textbf{Total} & \textbf{Delu.} & \textbf{Hallu.} & \textbf{Total} \\
\midrule
\multirow{9}{*}{\textit{Llama-3.1-8B-Instruct}}
& \texttt{10\%} & 21.7 & 43.9 & 16.6 & 45.4 & 70.0 & 52.5\\
& \texttt{20\%} & 43.8 & 72.2 & 33.0 & 75.8 & 92.9 & 78.9\\
& \texttt{30\%} & 54.5 & 81.6 & 41.2 & 74.2 & 91.8 & 76.0\\
& \texttt{40\%} & 69.1 & 89.2 & 53.1 & 79.2 & 93.0 & 79.3\\
& \texttt{50\%} & 72.7 & 91.3 & 56.9 & 76.7 & 92.7 & 78.3\\
& \texttt{60\%} & 79.4 & 93.6 & 65.8 & 79.8 & 93.3 & 80.7\\
& \texttt{70\%} & 87.0 & 96.0 & 74.9 & 86.3 & 95.1 & 85.4\\
& \texttt{80\%} & 93.2 & 98.5 & 86.4 & 90.6 & 97.0 & 91.4\\
& \texttt{90\%} & 96.2 & 99.1 & 93.4 & 93.3 & 99.1 & 94.8\\
\midrule
\multirow{9}{*}{\textit{Qwen2.5-7B-Instruct}}
& \texttt{10\%} & 28.7 & 46.8 & 21.8 & 61.9 & 83.0 & 65.0\\
& \texttt{20\%} & 42.3 & 65.6 & 32.4 & 75.9 & 93.2 & 78.7\\
& \texttt{30\%} & 50.2 & 75.8 & 39.4 & 82.3 & 96.0 & 84.2\\
& \texttt{40\%} & 63.3 & 86.7 & 49.9 & 83.9 & 96.8 & 85.8\\
& \texttt{50\%} & 69.5 & 90.6 & 56.1 & 84.2 & 96.6 & 85.2\\
& \texttt{60\%} & 76.3 & 93.1 & 62.0 & 84.9 & 96.2 & 84.9\\
& \texttt{70\%} & 81.1 & 95.6 & 68.4 & 84.7 & 96.4 & 84.6\\
& \texttt{80\%} & 87.6 & 97.2 & 75.8 & 86.4 & 97.0 & 86.9\\
& \texttt{90\%} & 95.0 & 98.9 & 87.9 & 93.6 & 99.2 & 93.9\\
\midrule
\multirow{9}{*}{\textit{Mistral-7B-Instruct-v0.1}}
& \texttt{10\%} & 21.1 & 41.9 & 25.9 & 36.6 & 63.0 & 50.0\\
& \texttt{20\%} & 42.1 & 63.1 & 46.1 & 61.2 & 81.4 & 69.7\\
& \texttt{30\%} & 61.0 & 76.3 & 61.4 & 73.1 & 89.1 & 80.2\\
& \texttt{40\%} & 74.5 & 85.2 & 73.4 & 79.7 & 93.0 & 86.1\\
& \texttt{50\%} & 81.7 & 89.3 & 80.2 & 84.4 & 94.7 & 89.2\\
& \texttt{60\%} & 86.5 & 92.3 & 85.1 & 87.7 & 95.5 & 91.3\\
& \texttt{70\%} & 91.1 & 94.5 & 89.2 & 90.4 & 96.6 & 93.3\\
& \texttt{80\%} & 94.4 & 96.8 & 92.9 & 91.6 & 97.0 & 94.5\\
& \texttt{90\%} & 97.2 & 98.0 & 96.1 & 94.0 & 98.2 & 96.5\\
\bottomrule
\end{tabular}
\caption{Complete results of refuse rate comparison with different SFT refuse data ratio.}
\label{table:by_ratio}
\end{table*}

\section{Prompts Used in Experiments}
\label{sec:Prompts}
\subsection{Prompts Used in Different Belief Estimation Methods}
\label{sec:Prompt_Belief}
The prompts are shown in Table~\ref{tab:prompts_for_belief_estimation}.

\begin{table*}[ht!]
\small
    \centering
    \begin{tabular}{m{\linewidth}<{\raggedright}}
        \toprule
        \rowcolor[gray]{0.95} 
        \textbf{Logits-based  Prompt} \\
        \midrule
        You are a helpful assistant. \\
        Answer the following question as accurately as possible. \\
        
        Question: \{question\} \\
        \midrule
        \rowcolor[gray]{0.95} 
        \textbf{P(true) Prompt} \\
        \midrule
        You are a helpful assistant. You should judge whether the answer to the given question is True or False. Please only reply with a simple word "True" or "False".\\
        Answer the following questions as accurately as possible. \\
        Question: \{question\} \\
        
        Answer: \{answer\} \\
        Is the above answer correct? (True / False) \\
        \midrule
        \rowcolor[gray]{0.95} 
        \textbf{Consistency Prompt} \\
        \midrule
        You are a helpful assistant. \\
        Answer the following question as accurately as possible. Provide ONLY the direct answer without any explanation. \\
        
        Question: \{question\} \\
        \midrule
        \rowcolor[gray]{0.95} 
        \textbf{Verb. 1S top1 Prompt} \\
        \midrule
        You are a helpful assistant, and you are always completely honest and DIRECT in your responses. \\
        Provide a brief, concise answer along with an explicit confidence percentage (0-100\%) about the correctness of your response. \\
        
        Question: \{question\} \\
        \midrule
        \rowcolor[gray]{0.95} 
        \textbf{Verb. 2S top1 Prompt} \\
        \midrule
        You are a helpful assistant, always completely honest and direct in your responses. You are also transparent about your confidence level and can honestly share how certain you are about the answer. \\
        
        Question: \{question\} \\
        Answer: \{previous\_answer\} \\
        How confident are you in the above answer (0-100\%)? \\
        \bottomrule
    \end{tabular}
    \caption{Prompts Used in Different Belief Estimation Methods.}
    \label{tab:prompts_for_belief_estimation}
\end{table*}

\subsection{Prompts of Different Honesty Levels}
\label{sec:Prompt_Reject}
The prompts are shown in Table~\ref{tab:prompts_honesty_levels}.
\begin{table*}[ht!]
\small
    \centering
    \begin{tabular}{m{\linewidth}<{\raggedright}}
        \toprule
        \rowcolor[gray]{0.95} 
        \textbf{"Helpful\_Can\_Refuse" Prompt} \\
        \midrule
            You are a helpful assistant. If you don't have information on the question, please answer with "I don't know". \\
            
            Answer the following question as accurately as possible. \\
            Question: \{question\} \\
        \midrule
        \rowcolor[gray]{0.95} 
        \textbf{"Helpful\_Less\_Refuse" Prompt} \\
        \midrule
            You are a helpful assistant. If you have relevant information to answer a question, please respond with your best and most accurate response. 
            Only refuse a question if you're completely uncertain or lack the necessary information to provide an answer. 
            In cases where you are unsure, or if the question falls outside your knowledge, simply reply with \texttt{``I don't know.''} 
            Try to answer questions to the best of your ability, and only reject a question when it is absolutely necessary, ensuring your answers are as helpful as possible. 
            Providing a response, even if not perfect, is often better than offering no answer at all. \\
            
            Answer the following question as accurately as possible. \\
            Question: \{question\} \\
        \midrule
        \rowcolor[gray]{0.95} 
        \textbf{"Helpful\_More\_Refuse" Prompt} \\
        \midrule
            You are a helpful assistant. If you lack sufficient information or confidence to answer a question correctly, simply reply with \texttt{``I don't know.''}
            It's important that your answers are accurate, and if you're uncertain, it's better to refuse the question than to provide incorrect information. 
            In such cases, you do not need to elaborate; just respond with \texttt{``I don't know.''}
            Prioritize accuracy over attempting to answer something you cannot be sure of. \\
            
            Answer the following question as accurately as possible. \\
            Question: \{question\} \\
        \midrule
        \rowcolor[gray]{0.95} 
        \textbf{"Helpful\_Medium\_Refuse" Prompt} \\
        \midrule
            You are a helpful assistant. If you have any doubt that your answer could be incorrect, please respond with \texttt{``I don't know.''}
            Accuracy is paramount, and if you're uncertain or lack sufficient information to confidently provide a correct response, refuse to answer. 
            Do not attempt to offer an answer if you are not confident. 
            Please prioritize certainty and precision in your responses. \\
            
            Answer the following question as accurately as possible. \\
            Question: \{question\} \\
        \midrule
        \rowcolor[gray]{0.95} 
        \textbf{"Helpful\_High\_Refuse" Prompt} \\
        \midrule
            You are a highly cautious and precise assistant. If you are uncertain or do not have enough information to provide a confident and accurate answer, you must refuse to answer. 
            Always prioritize caution and precision over attempting to answer a question. 
            If there is even the slightest doubt about the accuracy of your response, respond with \texttt{``I don't know.''}
            Never attempt to answer questions if you are not fully confident in your response. 
            If the question is unclear, or ambiguous, or you lack context, refuse to answer without hesitation. \\
            
            Answer the following question as accurately as possible. \\
            Question: \{question\} \\
        \midrule
        \rowcolor[gray]{0.95} 
        \textbf{"Helpful\_Most\_Refuse" Prompt} \\
        \midrule
            You are a highly cautious and precise assistant. If you are uncertain or do not have enough information to provide a confident and accurate answer, you must refuse to answer. 
            Always prioritize caution and precision over attempting to answer a question. 
            If there is even the slightest doubt about the accuracy of your response, respond with \texttt{``I don't know.''}
            Never attempt to answer questions if you are not fully confident in your response. 
            If the question is unclear, or ambiguous, or you lack context, refuse to answer without hesitation. \\
            
            Answer the following question as accurately as possible. \\
            Question: \{question\} \\
        \bottomrule
    \end{tabular}
    \caption{Prompts of Different Honesty Levels.}
    \label{tab:prompts_honesty_levels}
\end{table*}

\subsection{Prompts of Reflection}
\label{sec:Prompt_reflection}
The prompts are shown in Table~\ref{tab:prompt_reflection}.
\begin{table*}[ht!]
    \small
    \centering
    \begin{tabular}{m{\linewidth}<{\raggedright}}
    \toprule
    \rowcolor[gray]{0.95} 
    \textbf{Reflection on Previous Answer Prompt} \\
    \midrule
You are a helpful assistant. 
You will be given a question and your previous answer.
Please carefully reflect on whether your previous answer fully and accurately addresses the question. 
Consider if any part of your response could be misleading, incomplete, or inaccurate. 
If you identify any potential issues or areas for improvement, you should modify your answer accordingly. 
Only if, after thorough reflection, you believe your answer is completely accurate and there are no improvements to be made, should you respond with \texttt{``I insist.''} \\

Question: \{question\} \\
Previous Answer: \{previous\_answer\} \\
    \bottomrule
    \end{tabular}
    \caption{Prompt used to reflect on the previous answer's accuracy.}
    \label{tab:prompt_reflection}
\end{table*}

\subsection{Prompts of Retrieval-augmented Generation}
\label{sec:Prompt_rag}
The prompts are shown in Table~\ref{tab:prompt_rag}.
\begin{table*}[ht!]
    \small
    \centering
    \begin{tabular}{m{\linewidth}<{\raggedright}}
    \toprule
    \rowcolor[gray]{0.95} 
    \textbf{RAG Prompt} \\
    \midrule
You are a helpful assistant who provides the best possible answer. When uncertain about something, you'll make a guess.
Based on the provided documents, answer the following question. \\
Question: \{question\} \\
If the provided documents are not helpful,  you should answer based on your own knowledge.\\
Documents: \{passages\}\\
    \bottomrule
    \end{tabular}
    \caption{Prompt used in Retrieval-augmented Generation.}
    \label{tab:prompt_rag}
\end{table*}

\end{document}